\documentclass[12pt,a4paper]{article}
\usepackage{multirow}
\usepackage{rotating,multirow}
\usepackage{graphicx,geometry}
\usepackage{amsmath}
\usepackage{graphicx}
\usepackage[printonlyused,nohyperlinks]{acronym}
\usepackage{pythonhighlight}
\usepackage{tikz}
\usetikzlibrary{positioning}
\usepackage{booktabs}
\usepackage{url}

\title{Confusion Matrices and Accuracy Statistics for Binary Classifiers Using Unlabeled Data: \newline The Diagnostic Test Approach}

\author{Richard Evans\thanks{MCC Biostatistics Core, University of Minnesota. Email: evan0770@umn.edu. GitHub: revans011}}

\date{ \today}

\begin{document}

\maketitle

\begin{abstract}
Medical researchers have solved the problem of estimating the sensitivity and specificity of binary medical diagnostic tests without gold standard tests for comparison. That problem is the same as estimating confusion matrices for classifiers on unlabeled data. This article describes how to modify the diagnostic test solutions to estimate confusion matrices and accuracy statistics for supervised or unsupervised binary classifiers on unlabeled data.
\end{abstract}

\section{Introduction}
Sometimes it is important to know the accuracy of a classifier on unlabeled data. The labels may be delayed, as in consumer purchasing predictions, or obtaining the labels is cost prohibitive. The labels may not exist, as for some medical conditions, for which the true gold standard diagnostic test (a 100\% sensitive and 100\% specific classifier) would require subjects be euthanized and autopsied to obtain labels. 

Epidemiologists and biostatisticians have developed statistical methods for assessing 
the sensitivity ($Se$) and specificity ($Sp$) of diagnostic tests when gold standard comparison tests are unavailable. In data science terms, the diagnostic test assessment data are unlabeled. In this article, I describe how to modify those diagnostic test statistical methods to estimate confusion matrices and accuracy statistics for binary classifiers.

Hui and Walter (1980) \cite{hui}, showed that,  under mild conditions, using two diagnostic tests (neither are 100\% sensitive and 100\% specific) on two populations provided enough information to estimate the sensitivities and specificities of both tests and the unknown disease prevalences in both populations of test subjects. 

They used frequentist statistical methods. Branscum, Gardner, and Johnson (2005) \cite{branscum}, applied Bayesian methods to the Hui and Walter (1980) \cite{hui}, two-test, two-population model, and other diagnostic-test-accuracy models as well. The focus of this article is to modify the Branscum et al. (2005)\cite{branscum} approaches to estimate the accuracy of classifiers on unlabeled data, that is, datasets of just features.

The first model in this article is a two-classifier, one-unlabeled-dataset model. The method requires a classifier and a different, second classifier, and one-unlabeled dataset. The model as real-world applications when there is enough prior information about the classifiers or the prevalence (prevalence, $\pi$, is the parameter representing the unknown proportion of 1's in the features) to accurately estimate the model parameters (two sensitivities, two specificities, and the prevalence). 

The advantages of the Bayesian approach is that expressing uncertainty about the accuracy statistics (e.g., the standard deviation of $F1$) is much easier than with the frequentist statistical approach. Also, some classifier accuracy models aren't estimable in the frequentist sense, but are with Bayesian methods. Finally, Bayesian methods may sharpen estimates when datasets are small or there is good prior knowledge about the dataset prevalence or the accuracy of the classifiers. 

If prior information is weak, inferences about the parameters may be improved using two unlabeled datasets instead of one. The second model described in this article uses two classifiers and two unlabeled datasets to provide inferences for six parameters: two sensitivities, two specificities, and two prevalences.\cite{branscum}

Next, this article describes how to use the Markov chains, which are samples from the posterior distributions of sensitivity, specificity, and prevalence to construct confusion matrices and estimate accuracy statistics. The last section is an example using the the two-classifier, one-unlabeled-dataset model.

\section{The Classifiers}

Let $C_A$ be the primary binary classifier, and let $C_B$ be a secondary, different classifier. They may be supervised classifiers originally trained on labeled data, and now need their accuracy evaluated on an unlabeled dataset, or they can be unsupervised classifiers. 

For statistical reasons the two classifiers should work on different principles.\cite{branscum} For example, the classifiers could be a classification tree and a KNN, but shouldn't be two kinds of classification trees or two kinds of KNN. As a guideline, when the classifiers err, they should not err for the same reasons. 

In practice, the primary classifier is the one we are interested in. The secondary classifier could be one trained to act as a classifier for the purposes of the accuracy evaluation of the primary classifier.

For example, suppose $C_A$ is a binary classifier in production, and its accuracy needs re-evaluation, but obtaining labeled data for re-evaluation is expensive or difficult. Then, to use the methods in this article, simply create any secondary unsupervised classifier, different in methodology from the primary classifier, and use it.

Alternatively, $C_B$ could be a classifier, supervised and previously trained, or unsupervised, with known or unknown accuracy.

For the two-datasets model, we assume the sensitivities and specificities of the two tests are the same for both datasets. That is a reasonable assumption, as in practice we would expect an adequately trained  classifier to maintain its accuracy, at least initially, in production. The diagnostic test analogy is helpful: A diagnostic test (e.g., a home pregnancy test) is put into production after its accuracy is assessed, and we expect that accuracy not to change much with time or over subpopulations.

Note that the terms primary and secondary are used for explanatory purposes. The models themselves do not distinguish between the classifiers.

\section{The Datasets}
The data for these models are the binary prediction lists from two classifiers, predicted from datasets of features. The second model described in this article requires the classifiers predict from two datasets of features. The second dataset has the same features at the primary dataset. However, it can be a contrived dataset. (For that matter, both datasets can be contrived.) The two datasets should have different prevalences.\cite{hui}

\section{Two Models for Assessing Classifier Accuracy}

\subsection{Two classifiers assessed on one unlabeled dataset}
\label{section:one}

Let $C_A$ be the primary binary classifier and let $C_B$ be the secondary binary classifier.  

The predictions from the two classifiers on the same dataset of features can be arranged a 2x2 cross-classification table, where $y_1$ represents number of cases where both classifiers are $1$, $y_2$ represents the number of cases where $C_A$ is $1$ but $C_B$ is $0$, $y_3$ represents the number of cases where $C_A$ is $0$ but $C_B$ is $1$, and $y_4$ represents number of cases where both classifiers are $0$:

\medskip

\begin{center}
\makebox[\linewidth][c]{
\begin{tabular}{ccccc}
\multicolumn{2}{c}{}&\multicolumn{2}{c}{\textbf{$C_A$}}\\
\multicolumn{1}{c}{}&\multicolumn{1}{c}{\textbf{}}
&\multicolumn{1}{c}{\text{1}}
&\multicolumn{1}{c}{\text{0}}\\  
\multicolumn{1}{c}{\multirow{2}{*}{\rotatebox{0}{\textbf{$C_B$}}}}
&\text{1}  &y1 &y3\\
&\text{0}  &y2 &y4\\
\end{tabular}
}
\end{center}

\medskip

The table shows four pieces of data, but there are five parameters ($Se_A$, $Sp_A$, $Se_B$, $Sp_B$, and $\pi$). The additional information needed to estimate the parameters comes from their prior distributions.

The likelihood is,

\begin{equation*}
        y_1,y_2,y_3,y_4  \sim Multinomial\!\left(p_1,p_2,p_3,p_4,n\right), 
   \end{equation*}
   
where $n=\sum_{i=1}^{4} y_{i}$. That is, $n$ is the length of the prediction list. 

The probabilities are,

\begin{align*}
    p_1 =\pi  Se_A  Se_{B} +\left( 1-\pi \right)  \left(1-Sp_{A}\right)  \left(1-Sp_{B}\right) \\
    p_2 =\pi  Se_A  \left(1-Se_{B}\right) +\left( 1-\pi \right)  \left(1-Sp_{A}\right)  Sp_{B} \\
    p_3 =\pi  \left(1-Se_{A}\right)  Se_{B} +\left( 1-\pi \right)  Sp_{A}  \left(1-Sp_{B}\right) \\
    p_4 =\pi  \left(1-Se_{A}\right)  \left(1-Se_{B}\right) +\left( 1-\pi \right)  Sp_{A}  Sp_{B}.
    \end{align*}

The prior distributions are,

\begin{equation*}
\begin{aligned}
&Se_A \sim Beta(a_{Se_A},b_{Se_A}) \; \; Sp_A \sim Beta(a_{Sp_A},b_{Sp_A}) \\
&Se_B \sim Beta(a_{Se_B},b_{Se_B}) \; \;  Sp_B \sim Beta(a_{Sp_B},b_{Sp_B}) \\
 &\pi \sim Beta(a_{\pi},b_{\pi}). 
\end{aligned}
\end{equation*}

This model is useful when most of the priors are informative. For example, when previous experience on a particular kind of dataset allows for a tight Beta prior for $\pi$, or $C_B$ is a classifier with known $Se_B$ and $Sp_B$. The BetaBuster software \cite{branscum} is a convenient way of determining the $a$'s and $b$'s based on expert opinion.

\subsection{Two classifiers assessed on two unlabeled datasets}

This model requires a second dataset of features. Let $\alpha$ represent one dataset of features, and $\beta$, the other. Predictions from two classifiers on two datasets make four lists of predictions that can be arranged into two 2x2 tables, one for each dataset. Those tables provide eight pieces of information for six parameters ($Se_A$, $Sp_A$, $Se_B$, $Sp_B$, $\pi_\alpha, $and $\pi_\beta$), making this model identifiable, and less sensitive to small changes in the prior distributions compared to the two-classifier, one-unlabeled-dataset model.

The likelihood is,

\begin{align}
        y_{1\alpha},y_{2\alpha},y_{3\alpha},y_{4\alpha}  \sim Multinomial \left(p_{1\alpha},p_{2\alpha},p_{3\alpha}p_{4\alpha},n_{\alpha} \right), \label{eq:multi1}
 \end{align}
 
 and
 
 \begin{align}
        y_{1\beta},y_{2\beta},y_{3\beta},y_{4\beta}  \sim Multinomial \left(p_{1\beta},p_{2\beta},p_{3\beta}, p_{4\beta},n_{\beta} \right), \label{eq:multi2}
\end{align}

where $n_{\alpha}=\sum_{i=1}^{4} y_{i \alpha}$ and $n_{\beta}=\sum_{i=1}^{4} y_{i \beta}$.  \\

The probabilities for (\ref{eq:multi1}) are,

\begin{align*}
    p_{1\alpha} =\pi_{\alpha}  Se_A  Se_{B} +\left( 1-\pi_{\alpha} \right)  \left(1-Sp_{A}\right)  \left(1-Sp_{B}\right) \\
    p_{2\alpha} =\pi_{\alpha}  Se_A  \left(1-Se_{B}\right) +\left( 1-\pi_{\alpha} \right)  \left(1-Sp_{A}\right)  Sp_{B} \\
    p_{3\alpha} =\pi_{\alpha}  \left(1-Se_{A}\right)  Se_{B} +\left( 1-\pi_{\alpha} \right)  Sp_{A}  \left(1-Sp_{B}\right) \\
    p_{4\alpha} =\pi_{\alpha}  \left(1-Se_{A}\right)  \left(1-Se_{B}\right) +\left( 1-\pi_{\alpha}\right)  Sp_{A}  Sp_{B}.
    \end{align*}

The probabilities for (\ref{eq:multi2}) are,

\begin{align*}
    p_{1\beta} =\pi_{\beta}  Se_A  Se_{B} +\left( 1-\pi_{\beta} \right)  \left(1-Sp_{A}\right)  \left(1-Sp_{B}\right) \\
    p_{2\beta} =\pi_{\beta}  Se_A  \left(1-Se_{B}\right) +\left( 1-\pi_{\beta} \right)  \left(1-Sp_{A}\right)  Sp_{B} \\
    p_{3\beta} =\pi_{\beta}  \left(1-Se_{A}\right)  Se_{B} +\left( 1-\pi_{\beta} \right)  Sp_{A}  \left(1-Sp_{B}\right) \\
    p_{4\beta} =\pi_{\beta}  \left(1-Se_{A}\right)  \left(1-Se_{B}\right) +\left( 1-\pi_{\beta}\right)  Sp_{A}  Sp_{B}.
    \end{align*}

The prior distributions are,

\begin{equation*}
\begin{aligned}
&Se_A \sim Beta(a_{Se_A},b_{Se_A}) \; \;                             &Sp_A \sim Beta(a_{Sp_A},b_{Sp_A}) \\
&Se_B \sim Beta(a_{Se_B},b_{Se_B}) \; \;                             &Sp_B \sim Beta(a_{Sp_B},b_{Sp_B}) \\
&\pi_{\alpha} \sim Beta(a_{\pi_{\alpha}},b_{\pi_{\alpha}}) \; \; & \pi_{\beta} \sim Beta(a_{\pi_{\beta}},b_{\pi_{\beta}}) 
\end{aligned}
\end{equation*}

\section{The Bayesian Calculations}
The object of this article was to describe the models, not the Bayesian analysis of them, which is a standard Bayesian analysis. There are many algorithms and language libraries to ``turn the Bayesian crank" and produce Markov chains that are representative samples from the posterior distributions of $Se_A$, $Sp_A$, $Se_B$, $Sp_B$, $\pi_\alpha,$ and $\pi_\beta$.

The chains can be used to calculate means, medians, standard deviations, and inter-quartile ranges for the parameters. For example, the sample average of the Markov chains for $Se_A$ is a point estimate for the mean of $Se_A$. The sample standard deviation of the Markov chains for $Se_A$ is a measure of error for the mean of $Se_A$. 

In addition, the chains can be combined to construct samples from other posterior distributions, such as the posterior distributions $F1$. Those posterior chains can then be used to calculate sample averages, and so on.

\section{The Confusion Matrix}
The Markov chains can be used to estimate a confusion matrix. The confusion matrix is not the cross tabulation of $C_A$ and $C_B$. It is the confusion matrix for $C_A$. All that is required to construct the matrix are the point estimates of $Se_A$ and $Sp_A$ obtained from the Markov chains representing their posterior distributions. 

Let $\widehat{Se_A}$ and $\widehat{Sp_A}$ represent the averages of the posterior chains for the primary classifier obtained from the Bayesian analysis of either the one-unlabeled-dataset model or the two-unlabeled-datasets model. Then the confusion matrix is,

\begin{center}
\begin{tikzpicture}[
box/.style={draw,rectangle,minimum size=2cm,text width=1.5cm,align=left}]
\matrix (conmat) [row sep=.1cm,column sep=.1cm] {
\node (tpos) [box,
    label=left:\( \mathbf{1} \),
    label=above:\( \mathbf{1} \),
    ] {$\hspace{5mm}\widehat{Se_A}$};
&
\node (fneg) [box,
    label=above:\textbf{0},
    label=above right:\textbf{},
    label=right:\( \mathrm{} \)] {$\hspace{1mm}1-\widehat{Se_A}$};
\\
\node (fpos) [box,
    label=left:\( \mathbf{0} \),
    label=below left:\textbf{},
    label=below:] {$\hspace{1mm}1-\widehat{Sp_A}$};
&
\node (tneg) [box,
    label=right:\( \mathrm{} \),
    label=below:] {$\hspace{5mm}\widehat{Sp_A}$};
\\
};
\node [left=.05cm of conmat,text width=1.5cm,align=right] {\textbf{Actual \\ Value}};
\node [above=.05cm of conmat] {\textbf{Classifier A Prediction}};
\end{tikzpicture}
\end{center}

The cell entries are between 0 and 1, inclusive. If counts are preferred, multiply the $\widehat{Se_A}$ and $\widehat{Sp_A}$ by the length of either training set and round if necessary. Error bounds for $\widehat{Se_A}$ and $\widehat{Se_A}$ can be included in the confusion matrix using standard deviations calculated from the Markov chains. The same process can be used to construct a confusion matrix for $C_B$ if it is of interest.

Other classifier accuracy statistics can be calculated from the Markov chains. Classifier accuracy statistic formulas are often presented in terms of cell counts, but they can be reformulated in terms of sensitivity, specificity, and prevalence. 
Accuracy is $Se_A \pi + Sp_A \left(1-\pi \right)$. Recall is $Se_A$. $F1$ is,

\begin{equation}
F1=2 \cdot \frac{Se_A PPV_A }{Se_A+PPV_A}
\end{equation}
where 

\begin{equation}
PPV_A = \frac{Se_A \pi}{Se_A \pi+\left(1-Sp_A  \right) \left(1-\pi  \right)}
\end{equation}

The advantage of using the Bayesian machinery is that it is easy to calculate point estimates and uncertainty intervals about the accuracy statistics using the Markov chains for sensitivity, specificity, and prevalence. 

For example, for F1, make an F1 Markov chain by calculating F1 using each step of the Markov chains for sensitivity, specificity, and prevalence. That new chain is a sample from the posterior distribution of F1. The average of the chain is a point estimate for the mean of F1, and the standard deviation of the chain is an estimate of the standard deviation of F1.

\section{Example: Two Classifiers, One Unlabeled Dataset}

Suppose our we have trained an AdaBoost classifier ($C_A$) on labeled data. On labeled test data its $Se_A=0.93$ and its $Sp_A=0.89$. Now, several months later, we fear the data have drifted and need to verify $C_A$'s accuracy on unlabeled data. The first step is to use $C_A$ to make a list of binary predictions for the unlabeled dataset.

Now we need another classifier, $C_B$, to make another list of predictions. One option is to use an unsupervised classifier to make the prediction list and then proceed with the Bayesian analysis. Instead, we will train a support vector machine (SVM) on the original training set used for the AdaBoost classifier, and test it on the original test data. The reason is because sensitivities and specificities calculated from the original test data can be used to inform, with appropriate uncertainty, the prior distributions for the model used to determine sensitivities and specificities on the new, unlabeled data.

Suppose the SVM has $Se_B=0.83$ and $Sp_B=0.85$ on the original labeled test dataset. Suppose the two lists of predictions from the two classifiers on the new, unlabeled data can be organized as,

\medskip

\makebox[\linewidth][c]{
\begin{tabular}{ccccc}
\multicolumn{2}{c}{}&\multicolumn{2}{c}{\textbf{$C_A$}}\\
\multicolumn{1}{c}{}&\multicolumn{1}{c}{\textbf{}}
&\multicolumn{1}{c}{\text{1}}
&\multicolumn{1}{c}{\text{0}}\\  
\multicolumn{1}{c}{\multirow{2}{*}{\rotatebox{0}{\textbf{$C_B$}}}}
&\text{1}  &40 &7\\
&\text{0}  &3 &100\\
\end{tabular}
}

\medskip

Using the model specification in Section~\ref{section:one}, the likelihood is,

\begin{equation*}
        40,3,7,100  \sim Multinomial\!\left(p_1,p_2,p_3,p_4,150\right), 
   \end{equation*}
   
The probabilities are,

\begin{align*}
    p_1 =\pi  Se_A  Se_{B} +\left( 1-\pi \right)  \left(1-Sp_{A}\right)  \left(1-Sp_{B}\right) \\
    p_2 =\pi  Se_A  \left(1-Se_{B}\right) +\left( 1-\pi \right)  \left(1-Sp_{A}\right)  Sp_{B} \\
    p_3 =\pi  \left(1-Se_{A}\right)  Se_{B} +\left( 1-\pi \right)  Sp_{A}  \left(1-Sp_{B}\right) \\
    p_4 =\pi  \left(1-Se_{A}\right)  \left(1-Se_{B}\right) +\left( 1-\pi \right)  Sp_{A}  Sp_{B}.
    \end{align*}

The prior distributions are,

\begin{equation*}
\begin{aligned}
&Se_A \sim Beta(20,4) \; \; Sp_A \sim Beta(20,4) \\
&Se_B \sim Beta(20,4) \; \;  Sp_B \sim Beta(20,4) \\
 &\pi \sim Beta(1,1). 
\end{aligned}
\end{equation*}

The Beta prior parameters $(20,2)$ were chosen to put 60\% of the Beta's mass above $0.8$, which is consistent with the sensitivities and specificities from the original test data. We had no idea what the prevalence might be, and gave that a noninformative $Beta(1,1)$ distribution.

For the new, unlabeled dataset, the accuracies of the AdaBoost classifier, as well as for the SVM and prevalence, are summarized in the following table.

\medskip

\begin{tabular}{lrrrrr}
\toprule
{} &  Sensitivity A &  Specificity A &  Sensitivity B &  Specificity B &  Prevalence \\
\midrule
count &        900.000 &        900.000 &        900.000 &        900.000 &     900.000 \\
mean  &          0.898 &          0.956 &          0.920 &          0.936 &       0.296 \\
std   &          0.014 &          0.006 &          0.012 &          0.008 &       0.013 \\
min   &          0.848 &          0.935 &          0.877 &          0.908 &       0.257 \\
25\%   &          0.889 &          0.952 &          0.912 &          0.931 &       0.288 \\
50\%   &          0.899 &          0.956 &          0.921 &          0.937 &       0.296 \\
75\%   &          0.909 &          0.960 &          0.929 &          0.942 &       0.304 \\
max   &          0.938 &          0.974 &          0.950 &          0.957 &       0.348 \\
\bottomrule
\end{tabular}

\medskip
\medskip
The counts in the table are the lengths of the Markov chains after burn in and thinning. \\

Using the Markov chain means for point estimates, $\widehat{Se_A}=0.898$ and $\widehat{Sp_A}=0.956$, the confusion matrix for the AdaBoost classifier is, \\

\begin{tikzpicture}[
box/.style={draw,rectangle,minimum size=2cm,text width=1.5cm,align=left}]
\matrix (conmat) [row sep=.1cm,column sep=.1cm] {
\node (tpos) [box,
    label=left:\( \mathbf{1} \),
    label=above:\( \mathbf{1} \),
    ] {$\hspace{2mm}  0.898$};
&
\node (fneg) [box,
    label=above:\textbf{0},
    label=above right:\textbf{},
    label=right:\( \mathrm{} \)] {$\hspace{0mm} 1-0.898$};
\\
\node (fpos) [box,
    label=left:\( \mathbf{0} \),
    label=below left:\textbf{},
    label=below:] {$\hspace{0mm} 1-0.956$};
&
\node (tneg) [box,
    label=right:\( \mathrm{} \),
    label=below:] {$\hspace{2mm} 0.956$};
\\
};
\node [left=.05cm of conmat,text width=1.5cm,align=right] {\textbf{Actual \\ Value}};
\node [above=.05cm of conmat] {\textbf{Classifier A Prediction}};
\end{tikzpicture}

\medskip
Using the Markov chains for $Se_A$, $Sp_B$, and $\pi$, with the formulas for Accuracy, Recall, PPV, and F1, the accuracy summaries for the AdaBoost classifier are, \\

\medskip

\begin{tabular}{lrrrr}
\toprule
{} &  Accuracy &   Recall &      PPV &       F1 \\
\midrule
count &   900.000 &  900.000 &  900.000 &  900.000 \\
mean  &     0.939 &    0.898 &    0.788 &    0.839 \\
std   &     0.006 &    0.014 &    0.027 &    0.021 \\
min   &     0.917 &    0.848 &    0.707 &    0.772 \\
25\%   &     0.935 &    0.889 &    0.770 &    0.825 \\
50\%   &     0.939 &    0.899 &    0.790 &    0.841 \\
75\%   &     0.943 &    0.909 &    0.807 &    0.854 \\
max   &     0.956 &    0.938 &    0.864 &    0.899 \\
\bottomrule
\end{tabular}

\medskip

\medskip

This example is in a notebook at $github/revans011/classifier\_accuracy$

\section{Discussion}
I have described two Bayesian models that can be used for evaluating the accuracy of classifiers on unlabeled data. The particulars of Bayesian analyses  and examples for these models are available on GitHub, $github.com/revans011/classifier\_accuracy$.

\end{document}